%% file: main.tex
\documentclass{article}
\PassOptionsToPackage{numbers, compress}{natbib}

\usepackage[preprint]{neurips_data_2021}

\usepackage[utf8]{inputenc} 
\usepackage[T1]{fontenc}    
\usepackage{hyperref}       
\usepackage{url}            
\usepackage{booktabs}       
\usepackage{amsfonts}       
\usepackage{latexsym,amsmath,amssymb}
\usepackage{nicefrac}       
\usepackage{microtype}      
\usepackage{xcolor}         
\usepackage{graphicx}
\usepackage{caption}
\usepackage{subcaption}
\usepackage{accents} 
\usepackage{wrapfig}

\usepackage[plain]{fancyref}

\usepackage{thmtools,thm-restate}

\newcommand{\lonenorm }{$L_1$-norm}
\newcommand{\ltwonorm}{$L_2$-norm}
\newcommand{\lone }{$L_1$}

\newcommand{\imone}{$\operatorname{IM1}$}
\newcommand{\imtwo}{$\operatorname{IM2}$}
\newcommand{\gb}{GB}
\newcommand{\gl}{GL}
\newcommand{\gen}{GEN}
\newcommand{\FID}{$\operatorname{FID}_{256}$}
\newcommand{\TCV}{$\operatorname{TCV}$}
\newcommand{\LVS}{$\operatorname{LVS}$}

\definecolor{myRed}{RGB}{228, 26, 28}
\definecolor{myBlue}{RGB}{55, 126, 184}
\definecolor{myGreen}{RGB}{77, 175, 74}
\definecolor{myPurple}{RGB}{152, 78, 163}
\definecolor{myOrange}{RGB}{252, 124, 4}

\usepackage{xspace}

\makeatletter
\DeclareRobustCommand\onedot{\futurelet\@let@token\@onedot}
\def\@onedot{\ifx\@let@token.\else.\null\fi\xspace}

\def\eg{\emph{e.g}\onedot} 
\def\ie{\emph{i.e}\onedot}

\makeatother

\title{On Quantitative Evaluations of Counterfactuals}
\author{%
    Frederik Hvilsh\o j\\
    Computer Science\\
    Aarhus University\\
    Aarhus, Denmark\\
    \texttt{fhvilshoj@cs.au.dk}
    
    \And
    
    Alexandros Iosifidis\\
    Electrical and Computer Engineering\\
    Aarhus University\\
    Aarhus, Denmark\\
    \texttt{ai@ece.au.dk}
    
    \And
    
    Ira Assent\\
    Computer Science\\
    Aarhus University\\
    Aarhus, Denmark\\
    \texttt{ira@cs.au.dk}
}

\begin{document}

\maketitle

\begin{abstract}
    As counterfactual examples become increasingly popular for explaining decisions of deep learning models, it is essential to understand what properties quantitative evaluation metrics \emph{do} capture and equally important what they \emph{do not} capture.
    Currently, such understanding is lacking, potentially slowing down scientific progress.
    In this paper, we consolidate the work on evaluating visual counterfactual examples through an analysis and experiments.
    We find that while most metrics behave as intended for sufficiently simple datasets, some fail to tell the difference between good and bad counterfactuals when the complexity increases.
    We observe experimentally that metrics give good scores to tiny adversarial-like changes, wrongly identifying such changes as superior counterfactual examples.
    To mitigate this issue, we propose two new metrics, the Label Variation Score and the oracle score, which are both less vulnerable to such tiny changes. 
    We conclude that a proper quantitative evaluation of visual counterfactual examples should combine metrics to ensure that all aspects of good counterfactuals are quantified.
\end{abstract}

\section{Introduction}\label{sec:introduction}
With the increased popularity of machine learning applications, a need for understanding machine learning models arises.
Many methods have been proposed to explain predictions of machine learning models.
To name but a few, some are based on heatmaps that identify salient input features~\cite{lime, lrp, deeptaylor, Chang2019}, others rely on transparent surrogate models~\cite{Guidotti2019, self-explaining-neural-networks}, and some produce counterfactual examples~\cite{Wachter2017, Dhurandhar2019, Singla2019}.
In this work, we consider the latter group, focusing on the image domain.

Counterfactual examples identify specific changes to inputs, such that the predicted outcome of a machine learning model changes.
Such examples allow interactions with the model to gain insights into its behavior.
For example, surveillance images of candidates picked out for screening can be assessed for biases by identifying features to change for the system to ignore the candidates~\cite{goyal19a}.

Work on counterfactual explanations has become increasingly popular \cite{stepin2021}. 
For images, counterfactuals should convey realistic changes to the input that are minimal and necessary while being valid, sparse, and proximal \cite{mothilal2020explaining, Dhurandhar2018, VanLooveren2019, Rodriguez2021}. 
Various metrics have been proposed to quantify aspects of the quality of counterfactual examples.
However, most metrics are used in isolation to evaluate a method proposed in the corresponding paper and to compare the method to others that are evaluated on different metrics. 
Furthermore, there exists little or no research on what properties the different metrics actually capture.
Lacking standard metrics and knowledge about what metrics capture makes it difficult to compare methods, which potentially slows down scientific progress within the field. 

In this work, we analyze and evaluate existing metrics to understand what each metric expresses in terms of realistic and minimal changes.
Through experiments, we find that most metrics have the intended behavior for image datasets of lower complexity. 
For a more complex dataset, our experiments show how multiple existing metrics fail to distinguish between good and bad counterfactuals.
We also expose vulnerabilities of different metrics and propose to account for such vulnerabilities by reporting multiple scores in combination.
Counterfactual examples comprising tiny unrealistic changes are found to often yield unintended good scores.
To mitigate this issue, we propose two new metrics that are less susceptible to tiny changes and align well with qualitative evaluations.
We argue that when presenting a proper evaluation of a counterfactual method, 
the evaluation would need a metric for quantifying how realistic counterfactuals are, \eg, the Fr\'echet Inception Distance, and a metric like our Latent Variable Score to assert that the validity of the counterfactuals generalize. 
Upon publication, we will also publish an evaluation framework for easy comparisons of methods. 

\section{Counterfactual Examples}\label{sec:counterfactuals}
The counterfactual question seeks to find \emph{necessary} and \emph{minimal} changes to an input to obtain an alternative outcome from a classifier~\cite{Wachter2017}. 
An answer often comes in the form ``Had values $v_1, ..., v_m$ been $\hat v_1, ..., \hat v_m$ and all other values remained the same, then outcome Y would have been Z.'' 
For images, an answer would be a new image similar to the input but with specific features changed.

Naturally, counterfactual examples come in various forms and might not convey the information that we would expect.
For example, adversarial attacks, which adds imperceptible noise to inputs in order to change predictions~\cite{adversarial-attacks}, are of little or no relevance in terms of interpretability. 
As such, multiple additional criteria have been proposed that counterfactuals need to possess to be intuitive for humans.

\paragraph{Realistic changes.}
To be useful for humans, counterfactual examples should look realistic \cite{Dhurandhar2018, Schut2021}. 
The criterion has also been described as counterfactuals being likely to stem from the same data distribution as the training data \cite{Poyiadzi2020, Dhurandhar2019}.
In the image domain, realistic changes can be hard to quantify.
How do you, for example, distinguish an adversarial attack from a proper counterfactual, when the attack may be closer to the input in terms of, \eg, Euclidean distance?
Methods for quantifying how realistic counterfactuals are typically rely on a form of connectedness \cite{Mahajan2019, Pawelczyk2020} or on embedding spaces of deep learning models \cite{VanLooveren2019, fid}.
Although being important for consolidating the field of counterfactual explanations, we find experimentally that  metrics have unintended behaviors when quantifying how realistic tiny adversarial-like changes are.

\paragraph{Minimal changes.} 
For counterfactual examples to be more useful to humans, input features need to change minimally for the prediction to change~\cite{Wachter2017}. 
Associated properties are sparsity and proximity, which relates to changing only few features and changing features such that the counterfactual stays in the proximity of the input~\cite{mothilal2020explaining}.
When only few features are changed, the counterfactuals are said to be more interpretable~\cite{Wachter2017}. 
In high-dimensional domains like the image domain, quantifying minimal changes with, \eg, Euclidean distance may yield undesired results.
For example, we demonstrate with an experiment that tiny adversarial-like changes can be deemed better (smaller) compared to realistic changes that naturally need to change more pixels.
In turn, a method that performs well only on minimal changes may be producing unrealistic adversarial-like counterfactuals that are of little or no value.
For a good performance on minimal changes to be meaningful, a method must thus also perform well on metrics quantifying how realistic the counterfactuals look.

\paragraph{Additional properties.}
In an interactive setting, computational efficiency is important~\cite{VanderWaa2018, VanLooveren2019, ecinn}.
If computations are too slow, interactions with a system will be poor. 
Although computation time is important, we do not study it here, as it does not quantify the quality of the counterfactuals.
It is also important that humans can use the generated counterfactuals.
Consequently, multiple works have done human studies of their methods~\cite{goyal19a, Dhurandhar2018, Singla2019}.
Such tests are typically domain specific and thus prohibit a generalized test.
Therefor, we do not include them for further evaluation here. 

\section{Evaluating Counterfactuals Quantitatively}\label{sec:evaluating-counterfactuals}
In this section, we present those quantitative metrics which have been applied to images in at least two publications and analyze their applicability in terms of how they measure changes. 
To mitigate an observed issue with tiny adversarial-like changes, we additionally propose two new metrics.
We find that each metric reflects specific aspects of counterfactual quality and need to be reported in combination with other metrics to avoid isolated drawbacks of the metrics.

\subsection{Existing Metrics}
\paragraph{Simple distance metrics.}
A natural first approach to measuring changes between inputs and counterfactuals is to use metrics like \lone~and \ltwonorm s even though
such norms are known to work poorly on high-dimensional data like images~\cite{Kang2020}.
We include those metrics because they are present in objective functions for gradient based counterfactual methods~\cite{Wachter2017, Dhurandhar2018, VanLooveren2019} and are in turn a natural first choice for measuring minimal changes.
In our experiments, we include a hybrid metric denoted the elastic net distance (EN), which is defined as $EN(x, c) = \|x - c\|_1 + \| x - c \|_2$, where $x$ is the input and $c$ is the counterfactual. 
We use this metric because it combines the \lone~ and the \ltwonorm.

\paragraph{Target-class validity.}
The Target Class Validity (\TCV)~\cite{Mahajan2019} quantifies the percentage of the generated counterfactuals that are predicted to be of the target class by the classifier under consideration: 
\begin{equation}\label{eq:tcv}
    \operatorname{TCV} = \frac{1}{|X|} \sum_{x \in X} 1_{[ f(x) \neq f( cf(x) )]}. 
\end{equation}
In Equation \eqref{eq:tcv}, $1_{[\cdot]}$ is the indicator function, $X$ is the test set, $f$ is the predictive function, and $cf(\cdot)$ is the function that generates the counterfactual examples.

The score quantifies how effective a method is in creating counterfactual examples that successfully change the class.
It does not quantify the quality of the counterfactuals in terms of neither minimal nor realistic changes. 
In turn, it should be reported along with other metrics quantifying those properties.

\paragraph{IM1.}
\cite{VanLooveren2019} introduces the \imone~score, which employs auto-encoders to approximate how well counterfactual examples follow the training data distribution.
The score shows the ratio between how well the counterfactual example $c$ of target class $q$ can be reconstructed by an auto-encoder trained on data from the target class $AE_q$ and an auto-encoder $AE_p$ trained on the data of the input class $p$:
\begin{equation} 
    \label{eq:IM1} 
  \operatorname{IM1}\left( c \right) = 
        \frac{
            \left\|c-\mathrm{AE}_{q}(c)\right\|_{2}^{2}}
        {
            \left\|c-\mathrm{AE}_{p}(c)\right\|_{2}^{2}+\epsilon
        }.
\end{equation}
A lower value means that $c$ follows the distribution of the class $q$ better than that of class $p$~\cite{VanLooveren2019}.

As argued in the previous section, it is important to measure how well counterfactual examples follow the distribution of the training data.
The \imone~score is a valuable tool for assessing such property.
As the score is quantitative, it also allows comparing different methods across publications.
Furthermore, the score is somewhat established as a metric, as multiple papers report the score~\cite{VanLooveren2019, Mahajan2019, Schut2021}.

The \imone~score can, however, be deceiving.
We find experimentally that methods which make tiny changes to the input can get an undesired good score, presumably because tiny changes yields almost no error even if the changes are not preserved by the auto-encoders.
Some classes may also be easier to reconstruct than others, resulting in skewed scores for different target classes. 
One target class may simply yield a lower numerator in \Fref{eq:IM1} than another target class, just because one is easier to reconstruct than the other. 
Finally, we know of no publicly available pre-trained auto-encoders for computing the score.
When new auto-encoders need to be trained for each publication, results may not be comparable across publications.
Through experiments, we demonstrate the issue by showing how, \eg, differences in normalization yield incomparable scores.

\paragraph{IM2.}
The \imtwo~score is also introduced in \cite{VanLooveren2019}.
It utilizes the discrepancy between reconstructions made by a class specific auto-encoder $AE_q$ and an auto-encoder trained on the entire training set $AE$:
\begin{equation}\label{eq:IM2}
\operatorname{IM2}\left(c\right)=\frac{\left\|\mathrm{AE}_{q}\left(c\right)-\mathrm{AE}\left(c\right)\right\|_{2}^{2}}{\left\|c\right\|_{1}+\epsilon}.
\end{equation}
%
According to the authors, a low value of \imtwo~indicates an interpretable counterfactual because the counterfactual follows the distribution of the target class as well as the distribution of the whole data set.
The applicability of the score is however debatable. 
\citet{Schut2021} demonstrate that the \imtwo~score fails to identify out-of-sample images.
\citet{Mahajan2019} also argue that both \imone~and \imtwo~are better reported by displaying both the denominator and numerator of each score.
For both \imone~ and \imtwo, we further find experimentally that when the complexity of the dataset increases, the computed scores get close to statistically insignificant amongst three different counterfactual methods.
In turn, the two metrics may be best suited for datasets of lower complexity. 

\paragraph{Fr\'echet Inception Distance.}
The Fr\'echet Inception Distance (FID) is a metric used for evaluating generative models~\cite{fid}.
The metric compares how similar two datasets are by comparing statistics of embeddings from the Inception V3 network~\cite{inceptionv3}. 
For counterfactuals, the score has been used to evaluate how well counterfactuals align with the original dataset~\cite{Rodriguez2021, Singla2019}.
FID is defined from mean Inception embeddings $\mu_1$ and $\mu_2$, and covariance matrices $\Sigma_1$ and $\Sigma_2$ of the test set and associated counterfactuals, respectively:
\begin{equation}
    \operatorname{FID} = \|\mu_1 - \mu_2\|_2^2 + tr[ \Sigma_1 + \Sigma_2 ] - 2tr\left[\sqrt{ \Sigma_1 \Sigma_2 }\right].
\end{equation}
In this work, we consider images that are smaller ($64\times 64$ pixels) compared to inputs of the Inception V3 model ($299\times299$ pixels).
Consequently, we compute the score for a different network.
We use embeddings from the last hidden layer of a convolutional neural network, which is identical to the model being explained by the counterfactual methods. 
The last hidden layer has $256$ output neurons, so we denote the score \FID, to avoid any misconceptions. 
Although the score depends on the embedding network, we believe that our results will extend to the Inception V3 network. 

The score is a good fit for evaluating whether generated counterfactuals follow the distribution of the training data, as it is currently the standard metric for evaluating generative models.
It does, however, not take into account the relation between each specific input and its associated output.
As such, the metric could, \eg, be ``fooled'' by a high performing generative model producing realistic samples independent of the inputs. 
Consequently, the metric should be reported in combination with another metric which evaluates the validity of each counterfactual.
We also find in experiments that methods generating tiny changes to the input may be deemed of high quality; maybe because the tiny changes are either filtered out by the embedding network or do not affect summarizing statistics of the score.

\subsection{New Proposed Metrics.}
Through experiments, we find that tiny changes similar to adversarial attacks often yield undesirable good scores.
To mitigate this issue, we introduce two new metrics. 
Both metrics rely on the assumption that tiny adversarial-like counterfactuals are very model specific~\cite{Liu2017}.
Under this assumption, evaluating counterfactuals on other classifiers should be less susceptible to tiny changes and more effective if the changes are semantically correct. 

\paragraph{Label Variation Score.} 
For datasets where each data point is associated with multiple class labels, 
individual classifiers for each class label can give insights into how each class is affected by a counterfactual change. 
Naturally, the class targeted by the counterfactual should be affected, while unrelated classes should not. 
At a high level, we use individual classifiers for each class label as a proxy for how much the concept related to the given class has changed in the counterfactual image. 

We propose the Label Variation Score (\LVS) to monitor predicted outcomes over different class labels.
\LVS~computes average Jensen Shannon (JS) divergences, denoted $d_{js}$, between predictions on inputs and counterfactuals.
Let $o_l$ be an ``oracle'' trained on the class label $l$, which outputs a discrete probability distribution over the labels.
then, \LVS~is defined as 
\begin{equation}\label{eq:lvs}
   \operatorname{LVS}_l = \frac{1}{|x|} \sum_{x \in x} d_{js}\left[ o_l(x) || o_l(cf(x))\right].
\end{equation}

As the score is based on individual classifiers for each class label, the score should be affected less by adversarial attacks. 
Intuitively, non-related labels should not be affected by counterfactuals and thus have a low \LVS, while labels that correlate with the counterfactual label may co-vary and get a higher \LVS.
For example, if an image of a face without makeup is changed to one with makeup, the face in the counterfactual should be predicted to smile as much as before, but the prediction of ``wearing lipstick'' may follow the prediction of ``wearing makeup'' as lipstick is a subset of makeup.

\LVS~yields a rich picture of which features are changed by the counterfactuals, and it allows human judgement of which features are allowed to be changed, as with the makeup and lipstick example.
Using the score, it becomes easier for humans to detect biases in the predictive model by identifying features that are changed unintendedly. 
On the contrary, \LVS~has the drawback that it needs multiple class labels to be applicable.
Also, attributes that are not labeled will not be possible to monitor.
In our experiments, \LVS~yields scores that align well with human interpretation on two different datasets.
We further verify the underlying hypothesis described above by finding that more realistically looking counterfactuals get better scores than, \eg, examples with tiny adversarial-like changes.

\paragraph{The oracle score.}
For datasets where \LVS~is not applicable, we propose to use a simpler metric which is based on training an additional classifier -- the oracle -- that is used to classify the counterfactuals examples. 
The score is the percentage of counterfactuals $C$ that are classified to the target class by both the classifier being explained $f$ and the oracle $o$:
\begin{equation}\label{eq:oracle}
    \operatorname{Oracle} = \frac{1}{|C|}\sum_{c\in C} 1_{[ f(c) = o(c) ]}.
\end{equation}

The score is similar to \TCV, but it is intended to avoid giving good scores to tiny adversarial-like changes. 
The oracle score depends on the additional oracle, which could tell more about the~oracle than the predictive model itself.
For example, it may be that adversarial attacks working on $f$ also work on $o$, which would wrongly yield a good score for such attacks.
However, we find experimentally that the score gives better scores for realistic counterfactual examples than tiny adversarial-like changes.

\section{Experiments}\label{sec:experiments} 
In this section, we study the above described metrics for different types of counterfactual methods to characterize what properties different metrics capture. 
We demonstrate through multiple experiments that no metric can express all desirable properties, and thus they should be used in combination.

\paragraph{Methods.} 
Throughout the experimental section, we compare three different methods for producing counterfactual explanations. 
The methods were chosen to represent a spectrum of methods ranging from gradient based methods producing sparse but less realistic changes in one extreme to methods based on generative models generating more realistic but larger changes in the other extreme.

In one end of the spectrum, \citet{Wachter2017} present a gradient based method (denoted \gb).
Counterfactuals are generated through gradient descent on the input to minimize a loss-function comprising an \lonenorm~``distance'' term which encourages minimal and sparse changes and a squared ``prediction'' loss on the predicted label, which encourages valid counterfactual examples.\footnote{We do not normalize each feature by the median absolute deviation, as images have identical value ranges.}
Another method that lies in this end of the spectrum is \cite{Dhurandhar2019} which follows a similar loss-function as \cite{Wachter2017}, but with a more complex distance function.
It should be mentioned that both methods were originally introduced for tabular data.
We here study it in the image domain as a simple method that produce counterfactuals with minimal changes that are looking less realistic. 

At the other end of the spectrum, we include the method proposed by \citet{ecinn} as a representative for methods based on generative models (denoted \gen). 
The method is based on conditional invertible neural networks (INNs) which are generative models that can also do classification~\cite{ibinn}.
Counterfactual embeddings are found by correcting embeddings of inputs such that the predicted class provably change. 
Counterfactual examples are successively generated by inverting the embeddings with the INN.
We find the method from \cite{ecinn} to be the most extreme case in this end of the spectrum, compared to, \eg, \cite{Rodriguez2021, Singla2019}, because it uses the same neural network for both predictions and for generating counterfactuals.
In contrast, \cite{Rodriguez2021} and \cite{Singla2019} train surrogate generative models, which are used for sampling counterfactuals. 
We study the method as a more complex method which produces more realistic counterfactuals but with larger amounts of change. 

At the middle of the spectrum, methods use gradients to compute counterfactuals similar \cite{Wachter2017}, but where gradient optimizations are guided by derivatives of generative models or other more sophisticated loss terms to enhance the quality of the counterfactuals~\cite{Dhurandhar2018, VanLooveren2019}.
In our experiments, we use the method proposed by \citet{VanLooveren2019} as representative (denoted \gl).
The method uses embeddings from an auto-encoder to optimize a class-prototype loss. 
Such method should produce counterfactuals where both the visual quality and the amount of changes is in between \gb~and \gen.
However, we find that in most cases, the visual quality is on par with \gb~in practice. 

\paragraph{Experimental details.}
The methods from \cite{Wachter2017} and \cite{VanLooveren2019} were implemented using the \verb!alibi! framework,\footnote{\url{https://docs.seldon.io/projects/alibi} (v.0.5.9), default parameters. Apache License 2.0.} and \cite{ecinn} was adopted from the official code.\footnote{\url{https://github.com/fhvilshoj/ECINN}, default parameters. MIT license.} 
The former two methods are used to identify counterfactuals for the same ``vanilla'' convolutional neural network, identical to the one described in \cite{VanLooveren2019}. 
The latter method is based on a conditional INN as predictive model, identical to that of \cite{ecinn}.
In turn, the presented results may be contributed to differences in architectures and not methods as such.
However, the goal of the experiments is not to identify a superior method, but to demonstrate properties of metrics for evaluating counterfactual explanations on images.
We note that the method from \cite{ecinn} generates counterfactuals for all classes different from the input class, so throughout experiments, we choose one target class uniformly at random for each input.
Additional experimental details on, \eg, hyperparameters for training are provided in the supplementary material.

Throughout the experiments, we report mean scores over the entire test set and 95\% confidence intervals in parentheses.
Except from \TCV, we report scores only on valid counterfactual examples from the test sets, \ie, we do not include counterfactuals that did not change the predicted class.

\subsection{FakeMNIST}\label{sec:sanity-check}
\begin{figure}[t]
    \begin{subfigure}{0.49\textwidth}
        \centering
        \caption{Counterfactual examples with target class $q=0$. }
        \label{fig:fake_mnist}
        \includegraphics{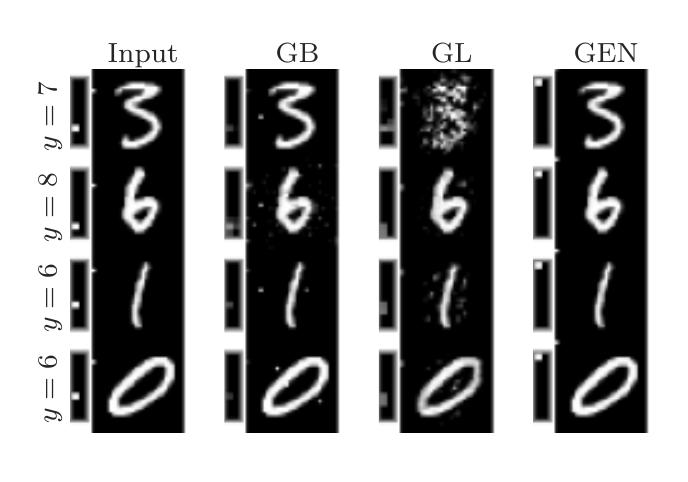}
    \end{subfigure}
    \begin{subfigure}{0.49\textwidth}
        \centering
        \caption{\LVS~scores.}
        \label{fig:fake-js}
        \includegraphics{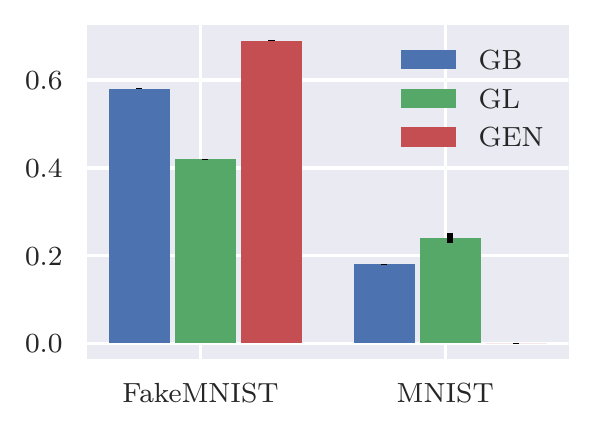}
    \end{subfigure}
    \caption{Experimental results for the FakeMNIST dataset~\cite{ecinn}.}
    \label{fig:fake-mnist-results}
\end{figure}

\cite{ecinn} propose FakeMNIST; an artificial dataset which dictates the relationship between pixels and labels.
To generate the dataset, MNIST images \cite{mnist} are shuffled and assigned new random labels.
The top-left $10\times 1$ pixels are colored according to the new labels, see first row of \Fref{fig:fake_mnist}.
The digits present in the images are independent of the labels while only the top-left pixels are label-dependent.
The dataset can be used to test whether counterfactual methods change only class-related features. 
There are, however, no metrics associated with the dataset.
We apply the \LVS~to further the evaluation protocol for the dataset and test if \LVS~detects methods that change label-independent features.

Column 1 of \Fref{fig:fake_mnist} displays four samples from the FakeMNIST test set.
Smaller rectangles magnify the top-left $10\times 2$ pixels for increased readability.
The first column displays inputs with labels 7, 8, 6, and 6, respectively (cf. labels or top-left dot locations).
The following three columns are counterfactuals with target class $q=0$, generated by the three representative methods.
In \Fref{fig:fake-js}, we show \LVS~for both the FakeMNIST labels and on the original MNIST labels.
As intended, the \LVS~finds that \gen~most successfully produces counterfactuals that change the predicted class (high \LVS~on FakeMNIST) and leaves the digit related pixels untouched (zero \LVS~on MNIST).
Furthermore, the \LVS~reveals that both \gb~and \gl~produces less effective counterfactuals, as their \LVS~on FakeMNIST are lower. 
Through the high \LVS~on MNIST, the metric also finds that the two methods wrongly alters digit related pixels when generating counterfactuals.  

In \Fref{tab:fake_mnist} in the supplementary material, we include scores of all other metrics described in \Fref{sec:evaluating-counterfactuals}.
All the scores behave as expected and quantify differences between the methods properly.
In conclusion, we find that for this simple dataset, qualitative observations and quantitative evaluations are well aligned in general.
In turn, we argue that to provide a complete picture of performance, new methods can provide all the presented scores for the FakeMNIST dataset.

\subsection{Normalization}\label{sec:normalization-matters}
Most metrics presented in this paper depend on data normalization or pretrained models.
The dependence makes reporting both data normalization and model specifications crucial for reproducibility.
We demonstrate the normalization issue with a practical example where we apply the metrics to the same counterfactuals but with different normalization.
The metrics have been adjusted to each normalization, \ie, new models were trained to operate on the particular normalization.

\input{tables/normalization-table.tex}

In \Fref{tab:mnist-scores}, we report mean scores for both a $[-0.5, 0.5]$ and a $[0, 1]$ normalization.  
By comparing the numbers between normalizations, we see that the best performing method for each metric is the same, independent of the normalization.
In \Fref{fig:models-avg-scores} in the supplementary material, we even find this result to be statistically significant across 10 independently initialized models. 
It should be noted that the $EN$ score is invariant to data shifts and scales linearly with the normalization range (cf. Table \ref{tab:mnist-scores}). 

As the table indicates, there is, however, an issue. 
Had the \imtwo~metric been used to compare \gl~with a $[-0.5, 0.5]$ normalization against \gen~with a $[0, 1]$ normalization, the conclusion would have been wrong, as \gl~would be deemed better than \gen.
Although this issue may seem obvious, it occurs in literature. 
If one compares reported \imtwo~scores between \cite{VanLooveren2019} and \cite{Mahajan2019}, the difference is about an order of magnitude.
\cite{VanLooveren2019} use normalization range $[-0.5, 0.5]$, while \cite{Mahajan2019} use $[0, 1]$.
We believe that the normalization differences contribute to explaining the difference between the reported scores.
In turn, we propose to establish a common set of models with a fixed normalization range to be used for every evaluation, such that comparison across publications becomes possible.
Upon publication, we will release our code to allow other researchers to easily evaluate their counterfactual methods.

\subsection{Inspecting Scores}\label{sec:extreme-inputs}
To get a deeper insight into how different metrics behave, we have identified pairs of inputs and counterfactuals for which there are unintended differences in scores.
We find that in some cases, which may be important for evaluating counterfactuals on specific datasets with specific properties, existing metrics can be a source of wrong conclusions when applied in isolation. 
Except for \FID, similar findings as those presented here were found for the CelebA-HQ dataset (see appendix).

\begin{figure}[b]
    \begin{subfigure}{0.32\textwidth}
        \centering
        \includegraphics{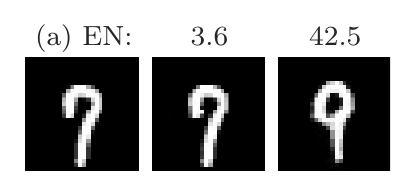}
    \end{subfigure}
    \begin{subfigure}{0.32\textwidth}
        \centering
        \includegraphics{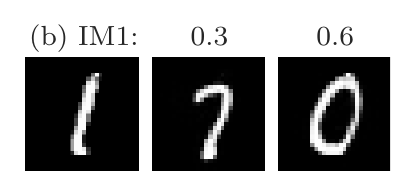}
    \end{subfigure}
    \begin{subfigure}{0.32\textwidth}
        \centering
        \includegraphics{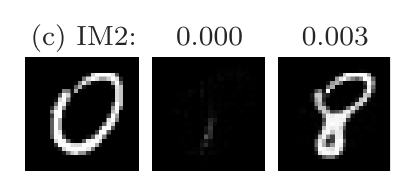}
    \end{subfigure}
    \caption{Examples of input (left) and counterfactual pairs (good: center, bad: right).}
    \label{fig:extremes}
\end{figure}

\paragraph{EN.}
For the image domain, the $EN$ distance is known to work poorly in terms of quantifying small interpretable changes \cite{Kang2020}. 
For completeness, we demonstrate the issue in \Fref{fig:extremes}a which shows a seven to the left and two counterfactuals with target class $q=9$ (center and right). 
The $EN$ distance is displayed above the two counterfactuals.
Arguably, the center image looks most like a seven and the right image looks like a nine.
However, according to the $EN$ distance, the center image is an order of magnitude better than the right. 
The example illustrates how tiny adversarial attacks may be deemed better than proper counterfactual examples, just because they change the input less.

\paragraph{IM1.}
\Fref{fig:extremes}b depicts a one with two counterfactuals with target class $q=7$ and $q=0$, respectively.
The \imone~score is meant to quantify how realistic counterfactual examples are.
Visually, the two counterfactual examples look similarly realistic.
The center image does, however, get twice as good a score compared to the right, \ie, the seven was deemed to be more realistic by the score.
Holding all else equal, this might be because more white pixels yield a larger loss.
For isolated cases like this, the \imone~score may produce undesirable results.
As observed in Table \ref{tab:mnist-scores}, the metric does, however, seem to work well on average when comparing methods on MNIST. 
As such, the \imone~score is best used as a summarizing statistic to compare averages of many samples across methods.

\paragraph{IM2.}
For the \imtwo~score, which should yield lower values for more interpretable counterfactuals, \Fref{fig:extremes}c shows how the score gives an almost completely black image a better score than an image of an eight digit.
On the contrary, the right image with the worst score seems more interpretable from a human perspective. 
The center image is presumably scoring best because it contains close to no information, which is easier to reconstruct than the right image which contains more information.
We demonstrate in the appendix that it holds more generally that the \imtwo~score decreases, when we decrease pixel values toward their minimal value.
In turn, the \imtwo~score might wrongly give good scores to methods that produce less interpretable counterfactuals by removing information from the inputs. 
To account for such drawback, also reporting, \eg, the oracle score, will make it harder to get good scores on both at once.
A high score on both metrics at once is thus preferred.

\paragraph{FID$_{256}$.}
\begin{wrapfigure}[21]{r}{0.23\textwidth}
    \centering
    \vspace{-2.3em}
    \includegraphics[width=0.75\linewidth]{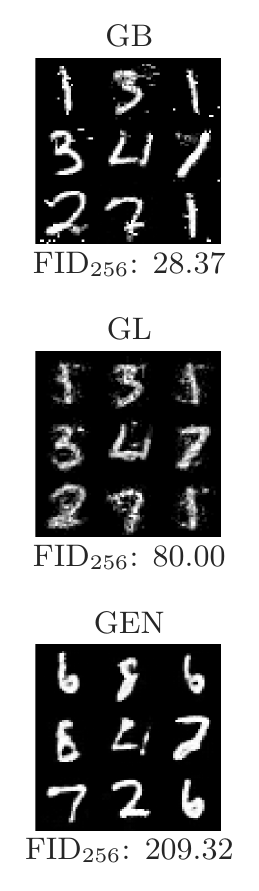}
    \caption{Counterfactuals on MNIST.}
    \label{fig:mnist-faed}
\end{wrapfigure}
For \FID, which quantifies the population wide similarity of sets of embedded images, it is not possible to identify single extreme samples. 
Instead, we observe how well \FID~distinguishes realistic and unrealistic samples. 
\Fref{fig:mnist-faed} shows counterfactual examples and the test-set-wide \FID~scores.
For a human observer, both \gb~and \gl~does a poor job in generating realistic changes.
It is, \eg, harder for humans to identify the target class for the two methods.
\FID~does not identify realistically looking samples in this particular case, as it yields better scores for both \gb~and \gl.
Interestingly, we find in the next section that \FID~successfully identifies the more realistically looking counterfactuals for the more complex CelebA-HQ dataset.
To deal with the identified issue, we argue that the \FID~score should be reported together with the \LVS~or the Oracle score, which are less vulnerable to tiny adversarial-like changes.
If both the \FID~and the \LVS~scores are good, then the quality of the counterfactuals is more likely to be high. 

In conclusion, we find that in isolated cases, the metrics may be deceiving.
We argue that the metrics should be reported jointly to account for each other's drawbacks.
For example, if a method gets a low \imtwo~score indicating interpretable counterfactuals, a low \FID~score indicating realistic counterfactuals, and a high oracle score indicating that the counterfactuals generalize, it is a strong indicator that it is a good method.

\subsection{Complex data} \label{sec:celeba}
In this section, we scale our experiments to the more complex dataset, CelebA-HQ~\cite{celeba}.
The goal is to evaluate how the studied metrics work in a more complex setting.

\begin{wrapfigure}[19]{r}{0.5\textwidth}
    \centering
    \vspace{-1em}
    \includegraphics{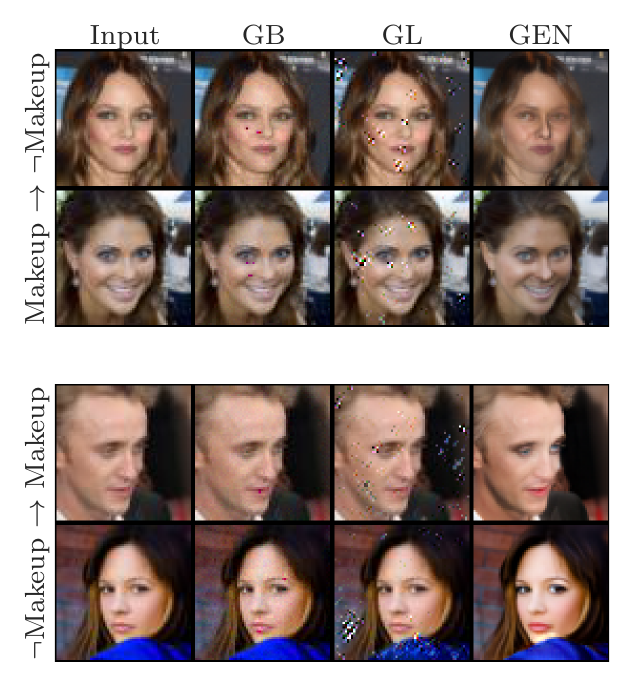}
    \caption{Counterfactuals for CelebA-HQ.}
    \label{fig:celeba}
\end{wrapfigure}

CelebA-HQ is a dataset of faces, where each sample is associated with 40 binary class labels.
\Fref{fig:celeba} presents four different inputs in the first column. 
The first two have a positive makeup label and the last two have a negative label.
The following three columns represent counterfactual examples of the opposite label value.
Qualitatively, we find the three compared methods to produce counterfactual examples with similar properties as for FakeMNIST and MNIST.
On the contrary, when we consider Table \ref{tab:celeba-scores}, we find that for some metrics the quantitative results vary from the previous experiments. 
Specifically, we see that the \imone~and \imtwo~scores fail to distinguish good from bad counterfactuals, as the scores yield almost the same value for all three methods.
We also observe that the \FID, in contrast to the previous experiment, successfully distinguishes the realistic from the unrealistic counterfactuals by giving \gen~the lowest (best) score and \gl~the highest.
In turn, for this more complex dataset, \FID~is not as vulnerable to tiny adversarial like attacks. 
For completeness, we also mention that the Oracle score and the $EN$ metric behave as expected. 
That is, the oracle score successfully identify the generative based method to most properly change the predicted class by the oracle, while the two other methods are found to be less successful.
The $EN$ metric correctly identifies the smallest changes, but the score is of little interest in the present comparison, as adversarial-like changes is still favored by the metric.
In a comparison of two methods which do not produce such tiny changes, the metric might, however, be valuable to quantify how much each method changes. 
\input{tables/celeba-table.tex}
\begin{wrapfigure}[9]{r}{0.5\textwidth}
    \centering
    \vspace{-1.5em}
    \includegraphics[width=\linewidth]{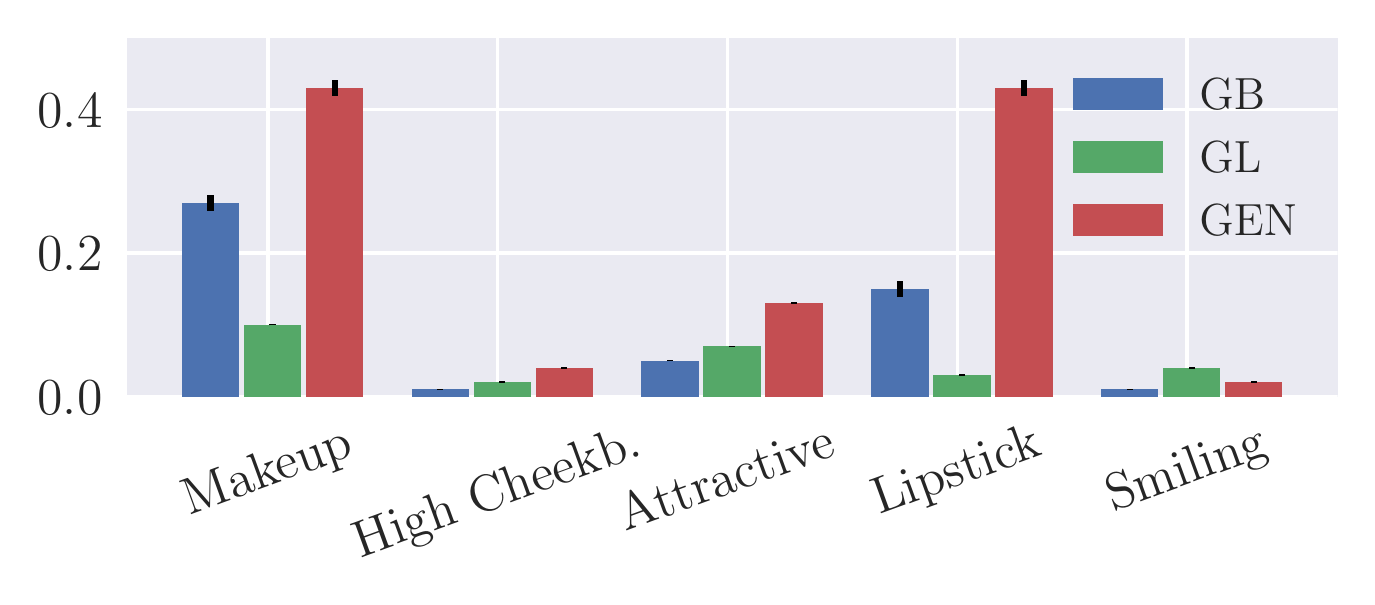}
    \caption{\LVS~for CelebA-HQ. Black vertical bars indicate 95\% confidence intervals.}
    \label{fig:js-divergences}
\end{wrapfigure}%
To also evaluate \LVS~on the more complex dataset, we have computed the score for the counterfactual label (smile versus no smile) and four other labels.
We chose the labels ``lipstick'' and ``attractive'' which should correlate more with the makeup label than the other two labels, ``high cheekbones'' and ``smiling.''
Also on this dataset, \LVS~successfully avoids giving the best stores for the tiny adversarial-like changes and favors more realistic changes. 
Specifically, \LVS~identifies that \gen~has a larger effect (high \LVS) for the related makeup, attractiveness, and lipstick labels, while having similar low effect (low \LVS) on the less related labels high cheekbones and smiling. 
\LVS~also successfully identifies how the changes made by, \eg, \gl~has less effect on all the labels, which indicates that the counterfactuals are highly model specific and behave more like adversarial examples. 

In summary, we find that for the more complex CelebA-HQ dataset, both the \imone~and the \imtwo~scores are less useful, while combining \FID~with the \LVS~yields a trustworthy quantitative evaluation of how realistic and valid the counterfactuals are, respectively.
Minimal changes are still hard to quantify, but with two methods that perform on par on \FID~and \LVS, the $EN$ distance may be applicable as to judge how much each method changes.

\section{Conclusion}
Through an analysis and experimental evaluations, we find that each quantitative metrics for evaluating visual counterfactual examples captures only some desired properties of good counterfactual examples.
On the sufficiently simple dataset FakeMNIST, we found that all metrics considered behaves as expected. 
However, on the more complex datasets like MNIST and CelebA-HQ, behaviors deviate more from the intended.
One particular issue is that visually unrealistic and tiny adversarial-like counterfactuals are very model specific and are often unintendedly deemed to be good by the metrics. 
To overcome this issue, we present the Label Variation Score and the oracle score, which are both based on surrogate predictive models that are less vulnerable to such tiny changes.
To make a proper quantitative evaluation of visual counterfactual examples, we conclude that capturing all the desired properties is best done by reporting metrics concerning both realistic changes and validity together.

\section{Limitations and Broader Impact}\label{sec:broader-impact}
By analyses and experimental evaluations of quantitative metrics for evaluating counterfactual examples, this work contributes to improve scientific progress within counterfactual examples.
As such, the work contributes to better understanding what can and can not be expected of different quantitative metrics.
Such understanding will arguably yield better evaluations of counterfactuals and consequently improve the performance of methods for generating counterfactual examples.
As such, we do not see any direct social impacts of this work.
Indirectly, improving counterfactual examples can potentially enable attackers to fool automated machine learning systems by creating realistically looking adversarial examples, which yield desired outcomes.

We also recognize the limitations of our work. 
First, by limiting our evaluation to metrics that have been published at least twice, we have not done a complete evaluation of all existing metrics for evaluating counterfactual examples.
In turn, there may be other metrics which better capture desired properties of counterfactual examples.
Second, to limit the scope, we have chosen three representative counterfactual methods which represents specific properties in counterfactuals.
As such, there may be other properties of counterfactual examples, that we have not evaluated and consequently do not know whether they effect metrics.
Finally, due to a large spread in datasets used across publications, we have restricted our evaluation to three datasets of increasing complexity.
From our work, it is not clear how our results extend to other datasets.

{ 
    \small  
    \bibliographystyle{plainnat}
    \bibliography{references.bib}
}

\appendix

\section{Additional Experimental Results}
\subsection{FakeMNIST}
In addition to the \LVS, we also ran all other metrics on the FakeMNIST dataset. 
\Fref{tab:fake_mnist} presents all the scores.
We find that all included scores behave as expected. 
Specifically, we expect the \imone~and \imtwo scores to identify that counterfactuals generated by \gen~are the most realistic, as they change only the top left pixels, which should be easier to capture by the auto-encoders compared to the more scattered changes by both \gb~and \gl. 
As only changing the top left pixels should produce a little difference in terms of the $EN$ distance, we would also expect \gen~to get the lowest score, which is also the case in \Fref{tab:fake_mnist}.
A similar argument also works for the \FID.
The \FID~should capture that the most realistic samples are those where only the top-left pixels are changed, which is also the case.

The \TCV is not based on the perceptual quality of the counterfactuals, but on how effective each method is in changing the predicted class on the given classifier.
We see from \Fref{tab:fake_mnist} that \gen~is the most effective, which aligns well with the rest of our experiments. 
Finally, we see that the oracle score, which indicates whether the counterfactual examples also generalize to another classifier, also identifies how counterfactuals from \gen~generalize better than those of \gb~and \gl. 

\begin{table}[t]
    \caption{
        Test set wide mean (95\% confidence intervals) on the FakeMNIST dataset. 
        Best scores are reported in bold.
     }
    \label{tab:fake_mnist}
    \centering
    \small
    \begin{tabular}{lcccccc}
        \toprule
         Method   & \TCV                     & $EN$                   & \imone                      &$100\cdot$\imtwo       & \FID               & Oracle                  \\
        \midrule                                                                                                         
         GB       & \hspace{\characterlength}68.11\% (0.91)           &  11.60 (0.49)          & 1.22 (0.01)                 & 0.49 (0.01)           & 252.5             & 88.31\% (0.76)           \\
         GL       & \hspace{\characterlength}84.07\% (0.72)           &  47.38 (0.90)          & 1.03 (0.00)                 & 1.23 (0.03)           & 309.95            & 55.51\% (1.06)           \\
         GEN      & \textbf{100.00\% (0.00)} &  \textbf{\hspace{\characterlength}6.81 (0.04)}  & \textbf{ 0.68 (0.00)}       & \textbf{0.21 (0.00)}  & \hspace{\characterlength}\hspace{\characterlength}\textbf{0.12}     & \textbf{99.98\% (0.03)}  \\
        \bottomrule
    \end{tabular}
\end{table}

\subsection{MNIST} 
\begin{figure}[b]
    \centering
    \includegraphics{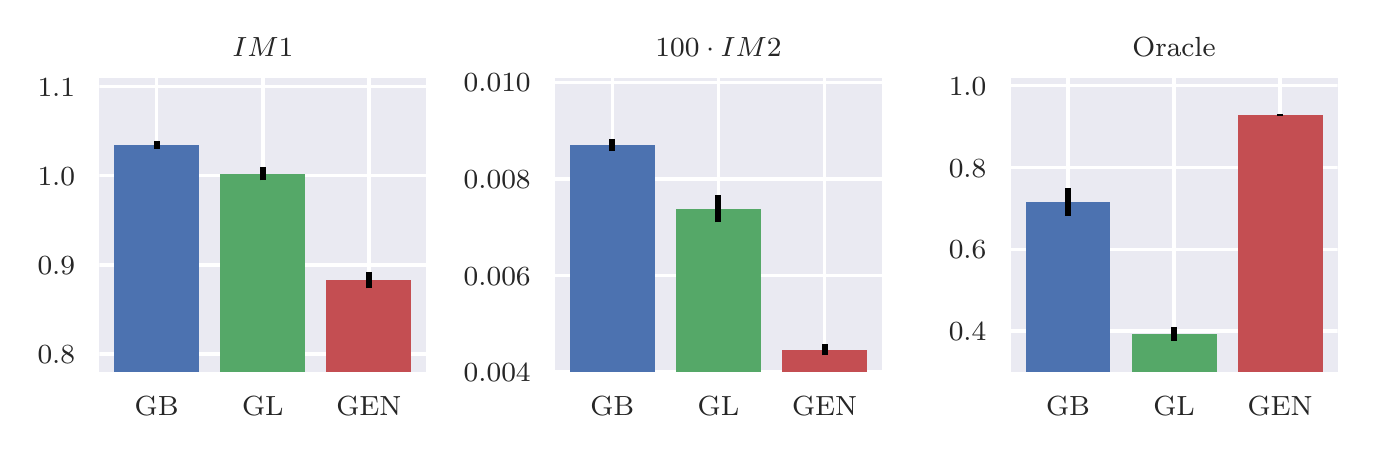}
    \caption{Mean scores on MNIST with 95\% confidence intervals for ten trials with ten randomly initialized evaluation models.}
    \label{fig:models-avg-scores}
\end{figure}

To evaluate how sensitive the model based scores \imone, \imtwo, and the oracle score are to initialization of models, we trained ten individual classifiers with different random initializations for the MNIST dataset and computed the mean scores along with 95\% confidence intervals.
In \Fref{fig:models-avg-scores}, we display the results, where bars represent mean values and horizontal black lined indicate confidence intervals.
From the figure, we see that all three scores have statistically significant differences on the 95\% level.
It should be mentioned, that we test ten identical model architectures. 
In turn, the experiment does not reveal any information on whether results are also robust across different model architectures. 

\subsection{Inspecting Scores on CelebA-HQ}
\begin{figure}
    \centering
    \begin{subfigure}{0.58\textwidth}
        \centering
        \caption{Counterfactuals examples for random CelebA-HQ sample.}
        \label{fig:celeba-extreme-samples}
        \includegraphics{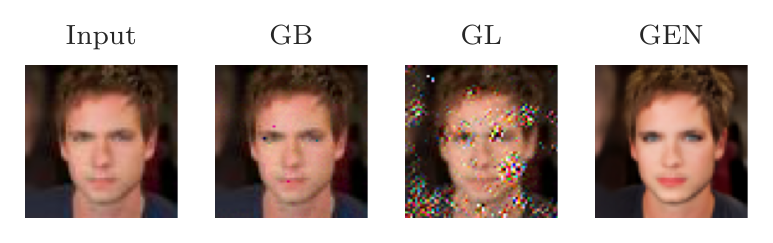}
    \end{subfigure}%
    \begin{subtable}{0.40\textwidth}
        \caption{Scores for counterfactuals in \Fref{fig:celeba-extremes}.\\}
        \label{tab:celeba-extreme-scores}
        \begin{tabular}{lccc}
            \toprule
             Method   &    $EN$ &  \imone &   $100\cdot$\imtwo \\
            \midrule
             GB       & \textbf{100.70} &  \textbf{1.00}    &  0.25 \\
             GL       & 787.69  &  \textbf{1.00}   &   0.47 \\
             GEN      & 569.36 &  1.12    &  \textbf{0.22} \\
            \bottomrule
        \end{tabular}
    \end{subtable}
    \caption{ An example of the behavior of the three scores $EN$, \imone, and \imtwo on counterfactuals adding makeup to a face. } 
    \label{fig:celeba-extremes}
\end{figure}

Similar to how we inspected scores on the MNIST dataset in \Fref{sec:extreme-inputs}, we have also considered similar input and counterfactual pairs for the more compled CelebA-HQ dataset.
Results are shown in \Fref{fig:celeba-extreme-samples}, which shows a random sample from the dataset along with the counterfactuals generated by the three counterfactual methods used in this work.
\Fref{tab:celeba-extreme-scores} shows the related scores. 
\Fref{fig:celeba-extremes} confirms the observations mentioned in \Fref{sec:extreme-inputs}, but also observations from our other experiment on CelebA-HQ (\Fref{sec:celeba}). 
Specifically, we see that $EN$ finds the tiny adversarial-like changes from \gb~to be the best, which does not align with what a human observer would deem a good counterfactual example.
As found in \Fref{sec:celeba}, \imone fail to distinguish counterfactuals from \gb~and \gl. 
Finally, the \imtwo~score yields similar scores for \gb~and \gen, is also in contradiction to the human observations, as the sample form \gen seems more interpretable. 

\section{Experimental Details}\label{sec:suppl-experimental-details}
In this section, we list all the relevant training details for the models used in this paper.
We note that we also supply code at \url{https://github.com/fhvilshoj/EvaluatingCounterfactuals}, which also contains all counterfactual examples used throughout the experiments, the code used for evaluation, and all the models used.

\paragraph{Convolutional Neural Networks.}
\gb~and \gl~both generate counterfactual examples for convolutaional neural networks with the model architecture described in \citet{VanLooveren2019}. 
For simplicity, we used the same model architecture for classifiers used with the oracle score, but with a different random initialization.
Unless explicitly stated differently in the main paper, all data was normalized to a $[-0.5; 0.5]$ range. 
All convolutional neural networks were trained with categorical cross-entropy. 
Remaining configurations for the convolutaional neural networks are stated in \Fref{tab:convolutional-configuration}.

\begin{table}[]
    \centering
    \caption{
        Training configurations for convolutional classifiers used throughout the paper.
        FakeMNIST and MNIST classifiers were trained in an identical manner, thus (Fake)MNIST means both FakeMNIST and MNIST.
        $^{\dagger}$ ADAM~\cite{adam} was used with Keras default parameters: $\beta_1=0.9, \beta_2=0.999, \epsilon=10^{-7}$.}
    \label{tab:convolutional-configuration}
    \begin{tabular}{lcccc}
        \toprule
        Configuration   & (Fake)MNIST classifier & (Fake)MNIST oracle & CelebA              \\
        \midrule
        Learning rate   & $10^{-3}$              & $10^{-3}$          & $10^{-3}$           \\
        Optimizer       & ADAM$^{\dagger}$       & ADAM$^{\dagger}$   & ADAM$^{\dagger}$    \\
        Batch size      & 64                     & 128                & 64                  \\
        epochs          & 10                     & 10                 & 100                 \\
       \bottomrule 
    \end{tabular}
\end{table}

\paragraph{Auto-encoders.} 
The auto-encoders used for generating counterfactuals for \gl~(\cite{VanLooveren2019}) and for computing \imone~and \imtwo~scores had the same architecture as described by \citet{VanLooveren2019}. 
We use independently initialized auto-encoders for computing and evaluating counterfactuals, respectively.
The models were trained with mean squared error loss and remaining configurations presented in \Fref{tab:auto-encoders}. 

\begin{table}[]
    \centering
    \caption{
        Training configurations for auto-encoders used throughout the paper.
        FakeMNIST and MNIST models were trained in an identical manner, thus (Fake)MNIST means both FakeMNIST and MNIST.
        $^{\dagger}$ Adam was used with Keras default parameters: $\beta_1=0.9, \beta_2=0.999, \epsilon=10^{-7}$.}
    \label{tab:auto-encoders}
    \begin{tabular}{lcc}
        Configuration   & (Fake)MNIST oracle & CelebA              \\
        \midrule
        Learning rate   & $10^{-3}$          & $10^{-3}$           \\
        Optimizer       & Adam$^{\dagger}$   & Adam$^{\dagger}$    \\
        Batch size      & 128                & 64                  \\
        epochs          & 50                 & 50                  \\
    \end{tabular}
\end{table}

\paragraph{Conditional INNs.}
The conditional INNs used in this paper used exactly the architectures and the loss function described in \cite{ibinn}.
We use $\beta=1.4265$ for FakeMNIST and MNIST and $\beta=1.0$ for CelebA, which was found to work well in \cite{ecinn}.
We only present the ``convincing'' counterfactuals from \cite{ecinn} with the $\alpha_1$-value suggested in the paper.
For both the FakeMNIST and MNIST datasets, we use the smaller architecture in \cite{ibinn} and for the CelebA dataset, we use the deeper architecture presented for the CIFAR10 dataset in \cite{ibinn}.
Additional configurations are presented in \Fref{tab:ecinn}.
We note that a full model specification and parameter configuration is also available in the public code repository.

\begin{table}[b]
    \centering
    \caption{
        Training configurations for auto-encoders used throughout the paper.
        All configurations are identical to those of \cite{ecinn}. 
        Stochastic Gradient Descent is abbreviated SGD below.}
    \label{tab:ecinn}
    \begin{tabular}{lcc}
        Configuration   & (Fake)MNIST oracle & CelebA              \\
        \midrule
        $\beta$         & 1.4265             & 1.0                 \\
        Learning rate   & 0.07               & $5\cdot 10^{-5}$    \\
        Optimizer       & SGD                & ADAM                \\
        Optimizer parameters & Momentum 0.9  & $\beta_1 = 0.95, \beta_2 = 0.99$\\
        Batch size      & 128                & 32                  \\
        epochs          & 60                 & 800                 \\
        Scheduler       & $10^{-1}$ milestone & $10^{-1}$ milestone \\
        Milestones      & 50                 & 200, 400, 600        \\
        Dequantization  & Uniform            & Uniform              \\
        Noise amplitude & $10^{-2}$          & $10^{-2}$            \\
        Label smoothing & $10^{-2}$          & 0                    \\
        Gradient norm clipping & 8.0         & 2.0                  \\
        Weight decay    & $10^{-4}$          & $10^{-4}$            \\
    \end{tabular}
\end{table}

\end{document}

%% file: tables/normalization-table.tex
\begin{table}[t]
    \centering
    \caption{Scores on MNIST for counterfactuals with different normalizations.}
    \label{tab:mnist-scores}
    \begin{tabular}{lccccc}
        \toprule
        Method  & $\operatorname{EN}$   & \imone                & $100 \cdot \operatorname{IM2}$    & \FID              & $\operatorname{Oracle}$   \\
        \midrule                                                                                     
        \multicolumn{6}{c}{$[-0.5;0.5]$ normalization}\\                                             
        \midrule                                                                                     
        \gb     & \textbf{16.07 (0.18)} & 0.99 (0.00)           & 0.55 (0.01)                       & \textbf{50.23}    & 73.38\% (0.87)\\
        \gl     & 42.76 (0.31)          & 0.99 (0.00)           & 0.53 (0.00)                       & 308.43            & 37.71\% (0.95)\\
        \gen    & 99.17 (0.58)          & \textbf{0.88 (0.00)}  & \textbf{0.17 (0.00)}              &  90.73            & \textbf{93.13\% (0.50)}\\
        \midrule
        \multicolumn{6}{c}{$[0;1]$ normalization}\\
        \midrule
        \gb &   \textbf{16.07 (0.18)}   &  1.06 (0.00)              & 2.46 (0.02)              &  \textbf{24.92}   & 48.25\% (0.98)          \\
        \gl &   42.76 (0.31)            &  1.04 (0.00)              & 1.94 (0.01)              & 173.82            & 38.53\% (0.95)          \\
        \gen &  99.17 (0.58)            &  \textbf{0.89 (0.00)}     & \textbf{1.47 (0.01)}     &  37.89            & \textbf{91.92\% (0.53)} \\
        \bottomrule
    \end{tabular}
\end{table}

%% file: tables/celeba-table.tex
\newlength{\characterlength}
\settowidth{\characterlength}{0}
\begin{table}[t]
    \centering
    \caption{CelebA-HQ scores}
    \label{tab:celeba-scores}
    \small
    \begin{tabular}{lcccccc}
        \\
        \toprule
         Method   & \TCV                    & $EN$                                              & \imone                 & $100\cdot$\imtwo     & \FID              & Oracle               \\
        \midrule                                                                                                                                 
         GB       & 96.07\% (0.72)          & \textbf{147.04 (\hspace{\characterlength}2.04)}   & \textbf{0.98 (0.00)}  & \textbf{0.47 (0.01)}  & 205.59            & 82.82\% (1.42)            \\
         GL       & 81.09\% (1.44)          & 344.02 (18.13)                                    & 0.99 (0.00)           & 0.52 (0.01)           & 484.08            & 32.84\% (1.92)            \\
         GEN      & \textbf{99.26\% (0.32)} & 684.26 (11.86)                                    & 1.03 (0.00)           & 0.53 (0.01)           & \textbf{98.35}    & \textbf{89.91\% (1.11)}   \\
        \bottomrule
    \end{tabular}
\end{table}